\documentclass[runningheads]{llncs}
\usepackage{graphicx}
\usepackage{url}

\usepackage{color}
\usepackage{verbatim}
\usepackage{indentfirst}
\usepackage[noadjust]{cite}
\usepackage{xcolor,graphicx}
\usepackage{float}
\usepackage{subfig}
\usepackage{amsmath}
\usepackage{makecell}
\usepackage{multirow, multicol}
\usepackage{threeparttable}
\usepackage{balance}
\usepackage{threeparttable}

\usepackage{amsmath} 

\makeatletter
\def\thickhline{%
	\noalign{\ifnum0=`}\fi\hrule \@height \thickarrayrulewidth \futurelet
	\reserved@a\@xthickhline}
\def\@xthickhline{\ifx\reserved@a\thickhline
	\vskip\doublerulesep
	\vskip-\thickarrayrulewidth
	\fi
	\ifnum0=`{\fi}}
\makeatother

\newlength{\thickarrayrulewidth}
\usepackage{booktabs}
\usepackage{color}
\usepackage{bbding}

\usepackage{graphicx}
\usepackage{amssymb}
\usepackage{xspace}

\usepackage{amsmath}
\usepackage{bm}

\begin{document}
\title{Categorical Relation-Preserving Contrastive Knowledge Distillation for Medical Image Classification}
%
%
\author{Xiaohan Xing$^1$, 
	Yuenan Hou$^2$, 
	Hang Li$^3$, 
	Yixuan Yuan$^4$, 
	Hongsheng Li$^1$, 
	Max Q.-H. Meng$^{1,5}$}
%
\authorrunning{X. Xing et al.}
\titlerunning{Categorical Relation-Preserving Contrastive Knowledge Distillation}
%
\institute{
	Department of Electronic Engineering, The Chinese University of Hong Kong, Shatin, Hong Kong, China \\
	\and
	Department of Information Engineering, The Chinese University of Hong Kong, Shatin, Hong Kong, China \\
	\and
	School of Informatics, Xiamen University, Xiamen, China \\
	\and
	Department of Electrical Engineering,
	City University of Hong Kong, Kowloon, Hong Kong, China \\
	\email{yxyuan.ee@cityu.edu.hk}
	\and
	Department of Electronic and Electrical Engineering, Southern University of Science and Technology, Shenzhen, China \\
	\email{max.meng@sustech.edu.cn}
}
%
\maketitle              
\begin{abstract}

The amount of medical images for training deep classification models is typically very scarce, making these deep models prone to overfit the training data. Studies showed that knowledge distillation (KD), especially the mean-teacher framework which is more robust to perturbations, can help mitigate the over-fitting effect. However, directly transferring KD from computer vision to medical image classification yields inferior performance as medical images suffer from higher intra-class variance and class imbalance. To address these issues, we propose a novel Categorical Relation-preserving Contrastive Knowledge Distillation (CRCKD) algorithm, which takes the commonly used mean-teacher model as the supervisor. Specifically, we propose a novel Class-guided Contrastive Distillation (CCD) module to pull closer positive image pairs from the same class in the teacher and student models, while pushing apart negative image pairs from different classes. With this regularization, the feature distribution of the student model shows higher intra-class similarity and inter-class variance.	Besides, we propose a Categorical Relation Preserving (CRP) loss to distill the teacher's relational knowledge in a robust and class-balanced manner. With the contribution of the CCD and CRP, our CRCKD algorithm can distill the relational knowledge more comprehensively. Extensive experiments on the HAM10000 and APTOS datasets demonstrate the superiority of the proposed CRCKD method. The source code is available at \textcolor{blue}{\url{https://github.com/hathawayxxh/CRCKD}}.

\keywords{Medical image classification  \and Knowledge distillation \and Contrastive learning.}
\end{abstract}
\section{Introduction}

With the recent progress of deep learning techniques, computer-aided diagnosis has shown human-level performance for some diseases and reduced the workload of human screening \cite{litjens2017survey}. 
However, the amount of training data for most diseases are limited, making the deep models prone to overfit the training data \cite{yang2019snapshot,zhuang2020deep}.
To tackle the over-fitting issue, many learning schemes have been proposed, such as transfer learning \cite{cheplygina2019not,shang2019leveraging}, dropout \cite{srivastava2014dropout}, and label-smoothing regularization \cite{szegedy2016rethinking,muller2019does}.
Another effective solution is knowledge distillation (KD), where a trained teacher model provides soft labels that supply secondary information to the student model, thus relieving the over-fitting problem \cite{hinton2015distilling}.

Among existing KD frameworks, the self-ensembling mean-teacher \cite{tarvainen2017mean} is widely studied in medical image classification.
Updated by the temporal moving average of the student model, the mean-teacher produces feature distribution and predictions that are robust to different perturbations, thus showing higher generalizability even with the limited amount of data.
Therefore, to train a student with high accuracy and generalizability, it is crucial to maximally distill knowledge from the mean-teacher.
Some researchers distilled the individual sample knowledge from the teacher, such as output logits \cite{thiagarajan2019distill} and feature maps \cite{wu2020leveraging}.
Recently, Liu et al. \cite{liu2020semi} took the relation among mini-batch samples as distilling targets and demonstrated its superiority over the individual KD counterparts.

However, most of the existing KD methods \cite{thiagarajan2019distill,wu2020leveraging,unnikrishnan2020semi,liu2020semi,abbasi2020classification,patra2019efficient,hou2019learning} are directly transferred from the computer vision field, without fully considering the following challenges in the medical domain.
First, the intra-class variation and inter-class similarity in medical datasets are more severe than those in the natural domain. 
In specific, two types of diseases may exhibit extremely similar color, shape, and texture, making them less distinguishable than two classes of natural images (dogs vs. cats).
Second, medical image datasets usually suffer from severe class imbalance since some diseases are common while others are rare.
Due to this, the knowledge distilled by current KD may be biased towards the majority class and has insufficient representation for the minority classes. 

To tackle the above-mentioned challenges, we propose a novel distillation approach, termed Categorical Relation-preserving Contrastive Knowledge Distillation (CRCKD), for medical image classification.
Built upon the mean-teacher framework, we propose two novel KD paradigms, i.e., \textit{Class-guided Contrastive Distillation} (CCD) and \textit{Categorical Relation Preserving} (CRP), to distill the rich structural knowledge from the mean-teacher model.
The main contributions are summarized as:
(1) We propose the CCD module to pull closer positive image pairs from the same class in the teacher and student models, while pushing apart negative image pairs from different classes. 
With this regularization, the feature distribution of the student model exhibits higher intra-class similarity and inter-class variance.
(2) To distill more robust and fine-grained relational knowledge, we propose the CRP loss that utilizes category centroids as anchors to regulate each sample's relation with different categories. 
Compared with previous relational KD \cite{tung2019similarity,liu2020semi} that adopts images in a mini-batch as anchors, the category centroids in our method serve as more reliable anchors and naturally mitigate the class imbalance problem.
(3) Experimental results on HAM10000 and APTOS datasets demonstrated the efficacy of our proposed CRCKD method, as well as the superiority of the CCD and CRP over existing relational KD paradigms.

\section{Method}

\begin{figure}[!h]
	\centering
	\includegraphics[width=0.95\textwidth]{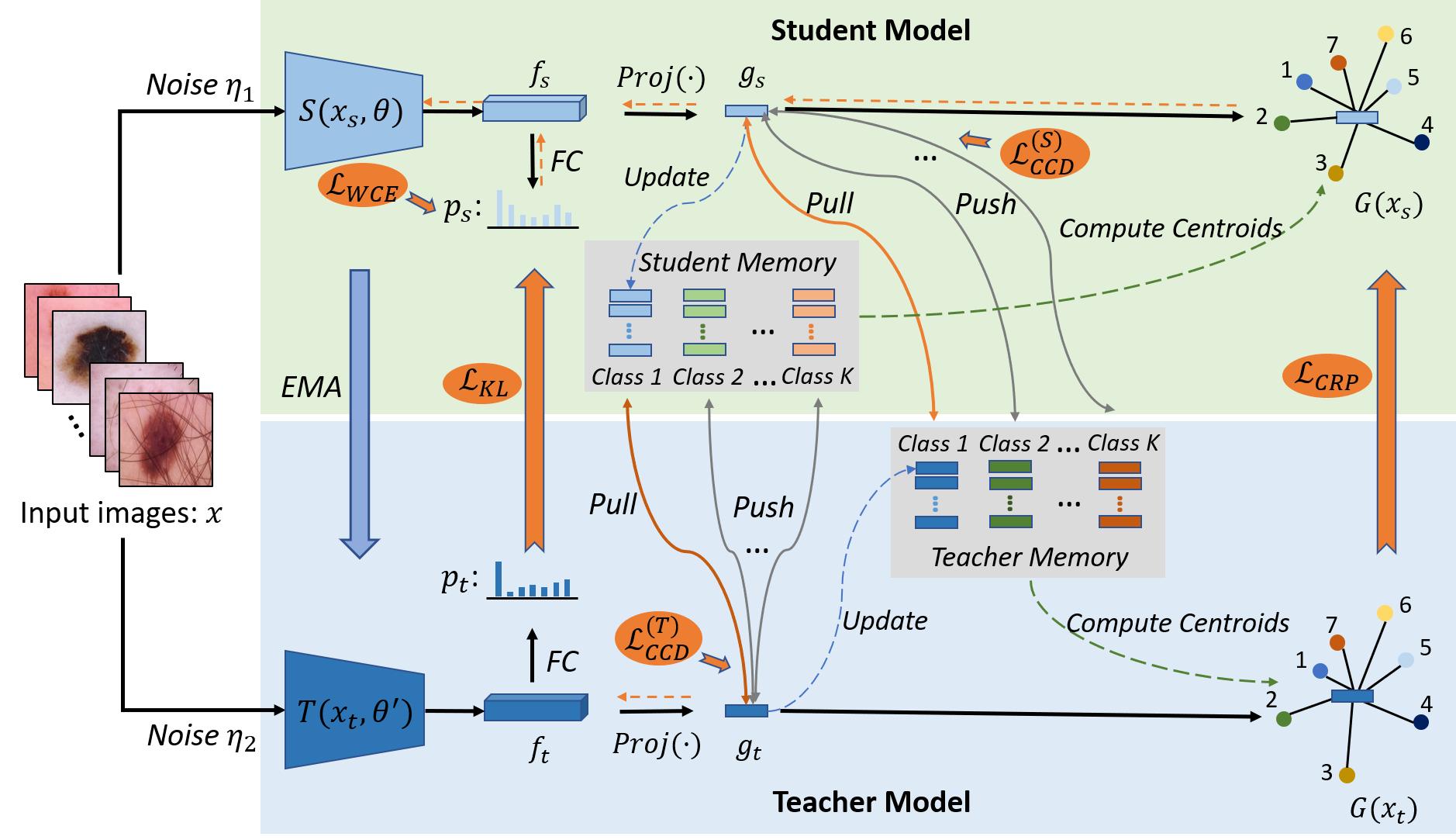}
	\caption{Overview of the proposed Categorical Relation-preserving Contrastive Knowledge Distillation (CRCKD) framework. The student model is supervised by the weighted cross-entropy loss $\mathcal{L}_{WCE}$ and knowledge distillation losses ($\mathcal{L}_{KL}$, $\mathcal{L}^{(S)}_{CCD}$, and $\mathcal{L}_{CRP}$). The dashed orange lines indicate the back-propagation paths of the gradients. [Best viewed in color]}
	\label{fig:overview}
\end{figure}

Fig. \ref{fig:overview} illustrates our proposed CRCKD framework. It consists of a student model and a mean-teacher model. 
The student model is optimized by stochastic gradient descent while the teacher weights $\theta^\prime$ are updated by the exponential moving average (EMA) of the student weights $\theta$.
Given an image $x$, it is augmented twice by adding different perturbations (i.e., random flipping and affine transformation) and produces two different images $x_s$ and $x_t$.
Taking the corresponding augmented image as an input, the student model and teacher model extract feature representations $f_s$ and $f_t$, and predict output probabilities $p_s$ and $p_t$, respectively. 
The student's prediction $p_s$ is supervised by the weighted cross-entropy loss $\mathcal{L}_{WCE}$ and the KL divergence $\mathcal{L}_{KL}$ with $p_t$. To constrain the consistency between the student's and teacher's structural information, we propose the $\mathcal{L}_{CCD}$ loss that pulls the positive pairs $(f_s, f_t)$ from the same class while pushing way the negative feature pairs from different classes.
Furthermore, in each model, we construct a novel relation graph $G(x_s)$ ($G(x_t)$) between the sample feature and the category centroids. 
The $\mathcal{L}_{CRP}$ loss is proposed to regularize the consistency between the relation graphs of the teacher and student models.

\subsection{Class-guided Contrastive Distillation (CCD)}

Recently, \textit{Contrastive Representation Distillation} (CRD) has achieved impressive distillation performance via incorporating contrastive learning into the conventional KD paradigm. Despite its appealing results, one major shortcoming of CRD is that it will mistakenly push apart images from the same class in the feature space, thus unavoidably enlarging the intra-class variance.

To tackle this dilemma, we propose a novel \textit{Class-guided Contrastive Distillation} (CCD) which utilizes the class-label information to guide the CRD.
Specifically, CCD regards samples from the same class as positive pairs and pulls their representations closer, while taking images from different classes as negative pairs and pushing their representations apart.
As depicted in Fig. \ref{fig:overview}, two different augmentations of an image are processed by the teacher and student models to generate feature embeddings $f_s$ and $f_t \in R^{D}$.
Then, the embeddings are projected to $g_s$ and $g_t \in R^{d} (d<D)$ through $Proj(f_s ; \zeta_s)$ and $Proj(f_t ; \zeta_t)$, where $\zeta_s$ and $\zeta_t$ denote the trainable parameters in the projection layers (which are instantiated as linear transformation).
The projected embeddings $g_s$ and $g_t$ are mapped to the unit hypersphere through $L_2$ normalization, thus their similarity can be measured by the inner product.
Inspired by \cite{tian2019contrastive}, for each sample $g_s$ in the student model, we define the CCD loss as
\begin{equation}
\mathcal{L}^{(S)}_{CCD}(\theta, \zeta_s) = -\frac{1}{k_P}\sum_{i=1}^{k_P}(\log \frac{e^{(g_s \cdot g_{t, i}/\tau)}}{e^{(g_s \cdot g_{t, i}/\tau)} + \frac{k_N}{M}} +\sum_{j=1}^{k_N}\log (1-\frac{e^{(g_s \cdot g_{t, j}/\tau)}}{e^{(g_s \cdot g_{t, j}/\tau)} + \frac{k_N}{M}})),
\label{eq:SCKD_s}
\end{equation}
where $\tau$ is the temperature that controls the concentration level, $k_P$ and $k_N$ denote the number of positive samples and negative samples, respectively.
$M$ is the cardinality of the dataset.
By minimizing $L^{(S)}_{CCD}$, the student model is optimized to produce feature representations that are more similar with the $k_P$ positive pairs while differing from the $k_N$ negative samples in the teacher model.
Similarly, the CCD loss for the teacher model is defined as
\begin{equation}
\mathcal{L}^{(T)}_{CCD}(\zeta_t) = -\frac{1}{k_P}\sum_{i=1}^{k_P}(\log \frac{e^{(g_t \cdot g_{s, i}/\tau)}}{e^{(g_t \cdot g_{s, i}/\tau)} + \frac{k_N}{M}} +\sum_{j=1}^{k_N}\log(1-\frac{e^{(g_t \cdot g_{s, j}/\tau)}}{e^{(g_t \cdot g_{s, j}/\tau)} + \frac{k_N}{M}})).
\label{eq:SCKD_t}
\end{equation}
It is noteworthy that the $\mathcal{L}^{(T)}_{CCD}$ loss merely updates the projection head of the teacher model.
The CCD loss regularizes the consistency of teacher and student's inter-sample structural knowledge by enlarging intra-class similarity and inter-class divergence between these two models, thus yielding performance gains.

As suggested in \cite{tian2019contrastive,saunshi2019theoretical}, a large number of negative samples is required to ensure the performance of contrastive learning. 
To get access to a large number of negative samples and avoid large batch size, we follow Wu et al.\cite{wu2018unsupervised} to construct a memory bank $M \in R^{N\times d}$ that stores the $d$-dimensional embeddings of all $N$ training images. We denote the memory bank for the student (teacher) model as $M_s$ ($M_t$).
As shown in Fig. \ref{fig:overview}, in each forward propagation, only the features of the query samples in the mini-batch are updated while all other samples retain their embeddings at the last step.
For each query sample $x_s$ in the student model, the $k_P$ positive and $k_N$ negative samples in Eq. \ref{eq:SCKD_s} are randomly selected from the teacher's memory $M_t$. 
Similarly, the Eq. \ref{eq:SCKD_t} is computed in a similar manner.

\subsection{Categorical Relation Preserving (CRP)}

Although the proposed CCD can regularize the structural consistency of the teacher and student, the regularization is relatively coarse since each sample pushes apart negative image pairs from different classes without differentiation.
However, some categories of diseases are much more similar than other categories, thus their distributions should be closer in the embedding space.
To capture fine-grained relational knowledge, \cite{park2019relational,tung2019similarity,liu2020semi} proposed to distill the pair-wise relations between data samples in a mini-batch. 
However, for the dataset with severe class imbalance, most samples in a mini-batch belong to the majority class, thus the constructed relation graphs may suffer from class bias.

To settle the above issues and capture class-balanced relational knowledge, we propose a novel \textit{Categorical Relation Preserving} (CRP) loss that utilizes category centroids to construct the relation graph.
Specifically, in the student (teacher) model, we compute the centroid of the $i$-th category by averaging the features of all samples in the $i$-th class (retrieved from the memory bank $M_s$ ($M_t$)):
\begin{equation}
C^S_i = \frac{1}{|C_i|}\sum_{m_s\in C(i)}m_s, \quad  C^T_i = \frac{1}{|C_i|}\sum_{m_t\in C(i)}m_t, 
\label{eq:class_center}
\end{equation}
where $|C_i|$ denotes the number of samples in the $i$-th class.
Then, for each query sample $x_s$ ($x_t$) in the mini-batch, we compute its cosine similarity with all category centroids in the student (teacher) model. 
After softmax over all classes, we obtain the categorical relation between the sample $x_s$ ($x_t$) and the $i$-th category:
\begin{equation}
R(x_s, C_i^S) = \frac{e^{g_s\cdot C^S_i}}{\sum_{i=1}^K e^{g_s\cdot C^S_i}}, \quad R(x_t, C_i^T) = \frac{e^{g_t\cdot C^T_i}}{\sum_{i=1}^K e^{g_t\cdot C^T_i}},
\label{eq:CCR_s}
\end{equation}
where $K$ is the total number of classes in the dataset, $g_s$ and $g_t$ denote the sample's representations extracted by the student and teacher, respectively.
Then, we propose the CRP loss to minimize the KL divergence between the teacher and student's categorical relation graphs:
\begin{equation}
\mathcal{L}_{CRP} = \sum_{x_s, x_t \in \mathcal{T}} \sum_{i=1}^{K} R(x_t, C_i^T) \log \frac{R(x_t, C_i^T)}{R(x_s, C_i^S)}.
\label{eq:CRP}
\end{equation}

Compared with existing relational KD \cite{park2019relational,tung2019similarity,liu2020semi}, using category centroids as anchors in our methods has two advantages:
1) One anchor (class centroid) is utilized to represent each class (regardless of the number of images in that class), thus the constructed relation graphs naturally mitigate the bias caused by class imbalance.
2) The anchors in the CRP retrieved from the memory bank (momentum aggregation of temporal steps) are more robust than the anchors that rely on the current step.
To this end, the CRP loss regularizes the reliable categorical relation graphs built with more robust and representative anchors, thus is expected to better mimic the relational knowledge in the teacher model.


\subsection{Training and Testing}

In the proposed framework, the weights $\theta$ of the student model and $\zeta_s, \zeta_t$ of the projection layers are optimized by the loss function defined as:
\begin{equation}
\mathcal{L} = \mathcal{L}_{WCE} + \lambda_1 \cdot \mathcal{L}_{KL} + \lambda_2 \cdot \mathcal{L}_{CCD} + \lambda_3 \cdot \mathcal{L}_{CRP},
\label{eq:loss}
\end{equation}
where $\mathcal{L}_{WCE}$ denotes the weighted cross-entropy loss supervised by the ground-truth labels. The weight for each class in $\mathcal{L}_{WCE}$ is inverse proportional to the number of images in that class. $\mathcal{L}_{KL}$, $\mathcal{L}_{CCD}$, $\mathcal{L}_{CRP}$ are used to distill the individual and structural knowledge from the teacher model. In the first $T$ epochs, the trade-off weights $\lambda_1$ and $\lambda_3$ would gradually ramp-up from $0$ to $1$ according to a Gaussian warming up function $\lambda(t) = 1 * e^{(-5(1-t/T)^2)}$, while $\lambda_2$ is set as $0.1$. After $T$ epochs, we fix the value of $\lambda_1$ and $\lambda_3$ as $1$, and set $\lambda_2$ as $0.01$.
The teacher weights $\theta^\prime$ are updated as the EMA of the student weights.

At the testing stage, we discard the mean teacher and the projection heads, so the inference time is the same as the vanilla student model.

\section{Experiments}

\subsection{Dataset and Implementation Details}

\subsubsection{Dataset:}
We evaluated our proposed CRCKD framework on the HAM10000 \cite{tschandl2018ham10000,codella2018skin} and APTOS datasets \cite{APTOS.org}. 
The HAM10000 consists of 10015 dermoscopy images labeled by 7 types of skin lesions.
In APTOS, there are 3662 fundus images for grading diabetic retinopathy into five categories.
These two datasets both suffer from severe class imbalance.
A detailed description of these two datasets is provided in the supplementary material.
For both datasets, we performed five-fold cross-validation 
and reported the average testing performance over the five folds.
We evaluated the classification performance by overall accuracy ($ACC$), average precision ($AP$), balanced multi-class accuracy ($BMA$), and $F1$ score.
Due to the class imbalance, $BMA$ is considered the most important metric in this task.


\subsubsection{Implementation:}
Our method was implemented in Python with the Pytorch library. We employed the pre-trained DenseNet121 \cite{huang2017densely} as the backbone of the teacher and student model.
The network was trained with two P40 GPUs in parallel and the batch size was set to 64. 
Adam with $\beta_1 = 0.5$ and $\beta_2 = 0.999$ was used for network optimization.
We trained the network for 80 epochs with ramp-up epoch $T$ set as 30.
The initial learning rate was set to $0.0001$ and decayed by the one-cycle schedule.
The temperature $\tau$ in Eq. \ref{eq:SCKD_s} and Eq. \ref{eq:SCKD_t} is set as $0.07$.
For each query sample, the number of positive pairs $k_P$ and negative pairs $k_N$ was empirically set as 20 and 4096, respectively.

\subsection{Experimental Results}

\setlength{\tabcolsep}{1.5mm}{
	\begin{table*}[t]
		\centering
		\caption{Five-fold cross-validation results on HAM10000 and APTOS datasets. The highest rankings are highlighted in \textbf{bold}. Our method: B2 + CCD + CRP. Detailed performance on the mean and standard deviation of each algorithm is provided in the supplementary material.}\label{tab1}
		
		\begin{tabular}{ c| c c c c| c c c c}
			\thickhline
			\multirow{2}*{Methods}  & \multicolumn{4}{c}{HAM10000} & \multicolumn{4}{|c}{APTOS} \\
			~ & ACC & AP &  BMA &  F1 & ACC & AP &  BMA &  F1 \\
			\hline	
			DenseNet121 (B1) & 84.30 & 74.16 & 72.19 & 72.53 & 83.83 & 71.85 & 67.51 & 69.14 \\
			B1 + MT (B2) & 85.01 & 74.19 & 76.07 & 74.38 & 83.77 & 71.66 & 68.79 & 69.89 \\
			B2 + CCD & 85.52 & 74.87 & 77.64 & 75.45 & 84.42 & 72.79 & 70.42 & 71.23 \\
			B2 + CRP & 85.32 & 75.06 & 77.06 & 75.37 & 84.47 & 73.07 & 69.87 & 71.07 \\
			\hline
			Our method &\textbf{85.66} & \textbf{76.35} & \textbf{78.07} & \textbf{76.45} &\textbf{84.87} & \textbf{73.18} & \textbf{71.90} & \textbf{72.22} \\
			\hline
			B2 + CRD \cite{tian2019contrastive} & 85.33 & 74.41 & 76.44 & 74.77 & 84.09 & 71.82 & 69.38 & 70.27 \\
			B2 + SP \cite{tung2019similarity} & 85.13 & 74.92 & 76.06 & 74.48 & 83.16 & 70.51 & 69.15 & 69.54 \\
			B2 + FitNet \cite{yim2017gift} & 84.13 & 72.85 & 76.38 & 73.98 & 83.76 & 71.97 & 69.88 & 70.52 \\
			\thickhline		
			
		\end{tabular}
		
\end{table*}}

\subsubsection{Quantitative Results:}
Table\ref{tab1} summarizes the performance of our CRCKD and baseline algorithms on the HAM10000 and APTOS datasets.
Compared with the vanilla student model ``DenseNet121 (B1)'', ``B1 + MT (B2)'' achieves much better classification performance, showing the superiority of introducing the mean-teacher guidance.
What's more, Table\ref{tab1} ($row \, 3-4$) indicates the effectiveness of the proposed CCD and CRP, because involving either of them leads to relatively better performance than ``B2''.
We conjecture that performance gains brought by CCD and CRP are attributed to the distillation of structural knowledge.
Further, the combination of CCD and CRP in our method achieves the best performance with $BMA$ of $78.07\%$ ($2.00\%$ higher than ``B2'') on the HAM10000 dataset and $BMA$ of $71.90\%$ ($3.11\%$ higher than ``B2'') on the APTOS dataset, demonstrating the efficacy of the proposed method.

To further validate the effectiveness of the proposed CCD and CRP, we performed a comparison with the other three KD paradigms (see the last three rows in Table\ref{tab1}). 
``B2 + CCD'' achieves better performance than ``B2 + CRD''\cite{tian2019contrastive}, indicating the contribution of introducing the class-label guidance, which is consistent with our analysis in Section 2.1.
Besides, the superiority over ``B2 + SP'' \cite{tung2019similarity} suggests that our proposed CRP can better distill relational knowledge by utilizing more robust and representative class centroids as anchors.
Finally, both ``B2 + CCD'' and ``B2 + CRP'' outperform ``B2 + FitNet'' \cite{yim2017gift} (a classical KD method that distills the intermediate features of individual samples), suggesting the necessity and superiority of relation-preserving KD studied in this work.

\begin{figure}[!t]
	\centering
	\includegraphics[width=0.95\textwidth]{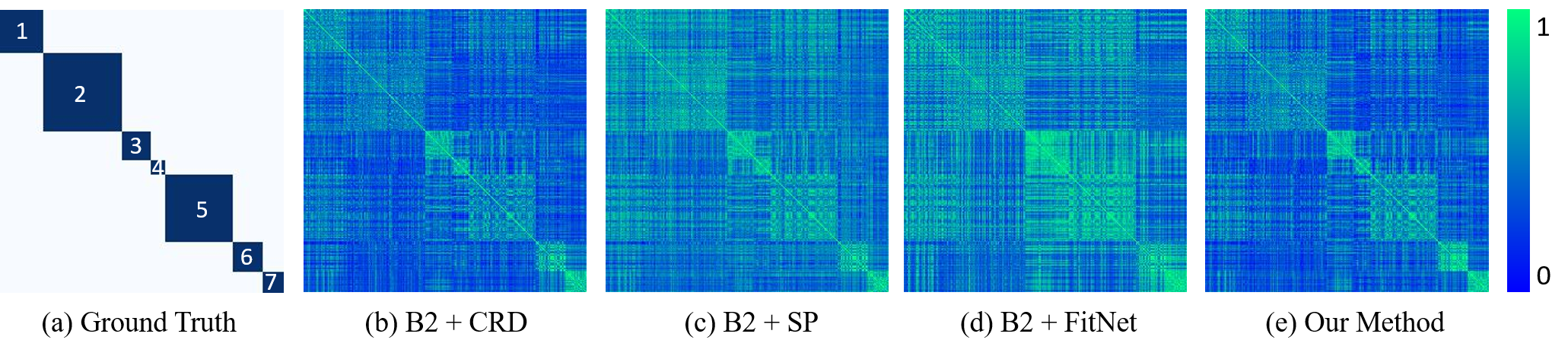}
	\caption{Visualization of the relation matrices between mini-batch samples (batch size=256) in the HAM10000 dataset. Input samples are grouped by ground truth class along each axis. 
		The color from dark blue to light green indicates increased similarity.
	}
	\label{fig:vis_relation}
\end{figure}

\subsubsection{Qualitative Analysis:}

Taking the HAM10000 dataset as an example, we visualize inter-sample relations produced by the features of different methods in Fig. \ref{fig:vis_relation}.
The diagonal blue blocks in Fig. \ref{fig:vis_relation} (a) denote intra-class similarity while other parts represent inter-class relation.
As shown in Fig. \ref{fig:vis_relation} (b), ``B2 + CRD'' exhibits low intra-class similarity (especially for the 1$st$, 2$nd$, and 5$th$ block).
In contrast, ``B2 + SP'' and ``B2 + FitNet''(Fig. \ref{fig:vis_relation} (c, d)) yield high inter-class relations.
With the regularization of the proposed CCD and CRP, our method (Fig. \ref{fig:vis_relation} (e)) exhibits lower inter-class similarity and higher intra-class similarity.
To quantitatively compare the relation matrices, we resort to $\mathcal{R}_{d} = \overline{\mathcal{R}}_{intra}/\overline{\mathcal{R}}_{inter}$, where $\overline{\mathcal{R}}_{intra}$ denotes the average pair-wise similarity between samples from the same class, $\overline{\mathcal{R}}_{inter}$ is the average sample similarity between different classes.
The larger value of $\mathcal{R}_{d}$ indicates higher intra-class similarity and lower inter-class similarity.
The $\mathcal{R}_{d}$ value of our method is $1.41$, outperforming the ``B2 + CRD'' (1.36), ``B2 + SP'' (1.35), and ``B2 + FitNet'' (1.22).
These results demonstrate that our proposed method can effectively alleviate the issue of high intra-class variance and inter-class similarity in the medical domain.

\setlength{\tabcolsep}{2.3mm}{
	\begin{table*}[t]
		\centering
		\caption{Comparison with state-of-the-art methods on the HAM10000 dataset.}\label{tab3}
		
		\begin{tabular}{ c| c c c c c}
			\hline
			Methods & ACC (\%) & AP (\%) &  BMA (\%) &  F1 (\%)\\
			\hline	
			Yan et al. \cite{yan2019melanoma} & 77.16$\pm$0.80 & 62.71$\pm$1.66 & 73.37$\pm$1.24 & 66.70$\pm$1.33  \\
			Zhang et al. \cite{zhang2018skin} & 82.34$\pm$0.73 & 72.32$\pm$1.38 & 74.01$\pm$1.76 & 72.28$\pm$1.04  \\		
			Zhang et al. \cite{zhang2019attention} & 81.61$\pm$0.94 & 71.44$\pm$1.22 & 73.34$\pm$1.86 & 71.65$\pm$0.85  \\		
			Liu et al. \cite{liu2020semi} & 84.73$\pm$1.00 & 73.88$\pm$1.24 & 76.55$\pm$1.32 & 74.63$\pm$0.99  \\
			\hline
			Our method & \textbf{85.66$\pm$0.97} & \textbf{76.35$\pm$0.99} & \textbf{78.07$\pm$1.28} & \textbf{76.45$\pm$0.66} \\
			\hline		
			
		\end{tabular}
		
\end{table*}}

\subsubsection{Comparison with Contemporary Methods:}

On the HAM10000 dataset, we further evaluated the performance of our proposed method with four state-of-the-art methods for skin lesion classification: attention-based methods \cite{yan2019melanoma,zhang2018skin}, synergic deep learning \cite{zhang2019attention}, and KD method based on sample relations \cite{liu2020semi}.
For a fair comparison, we changed \cite{liu2020semi} to full supervision, used the implementations of \cite{yan2019melanoma,zhang2018skin,zhang2019attention,liu2020semi} suggested by the authors, and evaluated the performance using the same dataset as our method.
As shown in Table \ref{tab3}, the proposed method outperforms existing methods with an improvement of $4.70\%$, $4.06\%$, $4.73\%$, $1.52\%$ in $BMA$, further validating the effectiveness of our proposed method.

\section{Conclusion}

In this paper, we present a novel Categorical Relation-preserving Contrastive Knowledge Distillation (CRCKD) framework for medical image classification.
Against the unique challenges of high inter-class similarity and class-imbalance in the medical domain, we propose two novel KD paradigms, i.e., CCD and CRP, to distill rich structural knowledge from the mean-teacher model.
Experimental results on the HAM10000 and APTOS datasets demonstrate the effectiveness of the proposed CCD and CRP over other KD paradigms.
On the HAM10000 dataset, experiments show that our CRCKD method outperforms many state-of-the-art methods.

\subsubsection{Acknowledgements.}
The work described in this paper was supported by National Key R\&D program of China with Grant No. 2019YFB1312400, Hong Kong RGC CRF grant C4063-18G, and Hong Kong RGC GRF grant \#14211420.

\bibliographystyle{splncs04}
\bibliography{paper545}

\begin{thebibliography}{10}
\providecommand{\url}[1]{\texttt{#1}}
\providecommand{\urlprefix}{URL }
\providecommand{\doi}[1]{https://doi.org/#1}

\bibitem{litjens2017survey}
Litjens, G., Kooi, T., Bejnordi, B.E., Setio, A.A.A., Ciompi, F., Ghafoorian,
  M., Van Der~Laak, J.A., Van~Ginneken, B., S{\'a}nchez, C.I.: A survey on deep
  learning in medical image analysis. Medical image analysis  \textbf{42},
  60--88 (2017)

\bibitem{yang2019snapshot}
Yang, C., Xie, L., Su, C., Yuille, A.L.: Snapshot distillation: Teacher-student
  optimization in one generation. In: Proc. CVPR. pp. 2859--2868 (2019)

\bibitem{zhuang2020deep}
Zhuang, J., Cai, J., Wang, R., Zhang, J., Zheng, W.S.: Deep knn for medical
  image classification. In: Proc. MICCAI. pp. 127--136. Springer (2020)

\bibitem{cheplygina2019not}
Cheplygina, V., de~Bruijne, M., Pluim, J.P.: Not-so-supervised: a survey of
  semi-supervised, multi-instance, and transfer learning in medical image
  analysis. Medical image analysis  \textbf{54},  280--296 (2019)

\bibitem{shang2019leveraging}
Shang, H., Sun, Z., Yang, W., Fu, X., Zheng, H., Chang, J., Huang, J.:
  Leveraging other datasets for medical imaging classification: evaluation of
  transfer, multi-task and semi-supervised learning. In: Proc. MICCAI. pp.
  431--439. Springer (2019)

\bibitem{srivastava2014dropout}
Srivastava, N., Hinton, G., Krizhevsky, A., Sutskever, I., Salakhutdinov, R.:
  Dropout: a simple way to prevent neural networks from overfitting. The
  journal of machine learning research  \textbf{15}(1),  1929--1958 (2014)

\bibitem{szegedy2016rethinking}
Szegedy, C., Vanhoucke, V., Ioffe, S., Shlens, J., Wojna, Z.: Rethinking the
  inception architecture for computer vision. In: Proc. CVPR. pp. 2818--2826
  (2016)

\bibitem{muller2019does}
M{\"u}ller, R., Kornblith, S., Hinton, G.: When does label smoothing help?
  arXiv preprint arXiv:1906.02629  (2019)

\bibitem{hinton2015distilling}
Hinton, G., Vinyals, O., Dean, J.: Distilling the knowledge in a neural
  network. arXiv preprint arXiv:1503.02531  (2015)

\bibitem{tarvainen2017mean}
Tarvainen, A., Valpola, H.: Mean teachers are better role models:
  Weight-averaged consistency targets improve semi-supervised deep learning
  results. In: Adv. Neural Inf. Process. Syst. pp. 1195--1204 (2017)

\bibitem{thiagarajan2019distill}
Thiagarajan, J.J., Kashyap, S., Karargyris, A.: Distill-to-label: Weakly
  supervised instance labeling using knowledge distillation. In: 2019 18th IEEE
  International Conference On Machine Learning And Applications (ICMLA). pp.
  902--907. IEEE (2019)

\bibitem{wu2020leveraging}
Wu, J., Yu, S., Chen, W., Ma, K., Fu, R., Liu, H., Di, X., Zheng, Y.:
  Leveraging undiagnosed data for glaucoma classification with teacher-student
  learning. In: Proc. MICCAI. pp. 731--740. Springer (2020)

\bibitem{liu2020semi}
Liu, Q., Yu, L., Luo, L., Dou, Q., Heng, P.A.: Semi-supervised medical image
  classification with relation-driven self-ensembling model. IEEE Trans. Med.
  Imaging  (2020)

\bibitem{unnikrishnan2020semi}
Unnikrishnan, B., Nguyen, C.M., Balaram, S., Foo, C.S., Krishnaswamy, P.:
  Semi-supervised classification of diagnostic radiographs with noteacher: A
  teacher that is not mean. In: Proc. MICCAI. pp. 624--634. Springer (2020)

\bibitem{abbasi2020classification}
Abbasi, S., Hajabdollahi, M., Khadivi, P., Karimi, N., Roshandel, R., Shirani,
  S., Samavi, S.: Classification of diabetic retinopathy using unlabeled data
  and knowledge distillation. arXiv preprint arXiv:2009.00982  (2020)

\bibitem{patra2019efficient}
Patra, A., Cai, Y., Chatelain, P., Sharma, H., Drukker, L., Papageorghiou,
  A.T., Noble, J.A.: Efficient ultrasound image analysis models with
  sonographer gaze assisted distillation. In: Proc. MICCAI. pp. 394--402.
  Springer (2019)

\bibitem{hou2019learning}
Hou, Y., Ma, Z., Liu, C., Loy, C.C.: Learning lightweight lane detection cnns
  by self attention distillation. In: Proc. ICCV. pp. 1013--1021 (2019)

\bibitem{tung2019similarity}
Tung, F., Mori, G.: Similarity-preserving knowledge distillation. In: Proc.
  ICCV. pp. 1365--1374 (2019)

\bibitem{tian2019contrastive}
Tian, Y., Krishnan, D., Isola, P.: Contrastive representation distillation.
  arXiv preprint arXiv:1910.10699  (2019)

\bibitem{saunshi2019theoretical}
Saunshi, N., Plevrakis, O., Arora, S., Khodak, M., Khandeparkar, H.: A
  theoretical analysis of contrastive unsupervised representation learning. In:
  International Conference on Machine Learning. pp. 5628--5637 (2019)

\bibitem{wu2018unsupervised}
Wu, Z., Xiong, Y., Yu, S.X., Lin, D.: Unsupervised feature learning via
  non-parametric instance discrimination. In: Proc. CVPR. pp. 3733--3742 (2018)

\bibitem{park2019relational}
Park, W., Kim, D., Lu, Y., Cho, M.: Relational knowledge distillation. In:
  Proc. CVPR. pp. 3967--3976 (2019)

\bibitem{tschandl2018ham10000}
Tschandl, P., Rosendahl, C., Kittler, H.: The ham10000 dataset, a large
  collection of multi-source dermatoscopic images of common pigmented skin
  lesions. Scientific data  \textbf{5},  180161 (2018)

\bibitem{codella2018skin}
Codella, N.C., Gutman, D., Celebi, M.E., Helba, B., Marchetti, M.A., Dusza,
  S.W., Kalloo, A., Liopyris, K., Mishra, N., Kittler, H., Halpern, A.: Skin
  lesion analysis toward melanoma detection: A challenge at the 2017
  international symposium on biomedical imaging (isbi), hosted by the
  international skin imaging collaboration (isic). In: Proc. ISBI. pp.
  168--172. IEEE (2018)

\bibitem{APTOS.org}
Aptos 2019 blindness detection.
  \url{https://www.kaggle.com/c/aptos2019-blindness-detection/data}

\bibitem{huang2017densely}
Huang, G., Liu, Z., Van Der~Maaten, L., Weinberger, K.Q.: Densely connected
  convolutional networks. In: Proc. CVPR. pp. 4700--4708 (2017)

\bibitem{yim2017gift}
Yim, J., Joo, D., Bae, J., Kim, J.: A gift from knowledge distillation: Fast
  optimization, network minimization and transfer learning. In: Proc. CVPR. pp.
  4133--4141 (2017)

\bibitem{yan2019melanoma}
Yan, Y., Kawahara, J., Hamarneh, G.: Melanoma recognition via visual attention.
  In: Inf Process Med Imaging. pp. 793--804. Springer (2019)

\bibitem{zhang2018skin}
Zhang, J., Xie, Y., Wu, Q., Xia, Y.: Skin lesion classification in dermoscopy
  images using synergic deep learning. In: Proc. MICCAI. pp. 12--20. Springer
  (2018)

\bibitem{zhang2019attention}
Zhang, J., Xie, Y., Xia, Y., Shen, C.: Attention residual learning for skin
  lesion classification. IEEE Trans. Med. Imaging  \textbf{38}(9),  2092--2103
  (2019)

\end{thebibliography}

\end{document}